\documentclass[lettersize,journal]{IEEEtran}
\usepackage{amsmath,amsfonts}
\usepackage{array}
\usepackage[caption=false,font=normalsize,labelfont=sf,textfont=sf]{subfig}
\usepackage{textcomp}
\usepackage{stfloats}
\usepackage{url}
\usepackage{verbatim}
\usepackage{graphicx}
\usepackage{cite}
\usepackage{booktabs}
\usepackage{multirow}
\usepackage{microtype}
\usepackage{mathtools}
\usepackage{amssymb}
\usepackage{amsthm}
\usepackage{booktabs}
\usepackage{bm}
\usepackage{textcomp}
\usepackage{url}
\usepackage{verbatim}
\usepackage{graphicx}
\usepackage{xcolor}
\usepackage{amsmath,amsfonts}
\usepackage{algorithm,algpseudocode}
\usepackage{booktabs}
\usepackage{hyperref}

\usepackage{array}
\theoremstyle{plain}
\newtheorem{theorem}{Theorem}
\newtheorem{proposition}{Proposition}

\newtheorem{corollary}[theorem]{Corollary}
\theoremstyle{definition}

\newtheorem{assumption}{Assumption}
\theoremstyle{remark}

\begin{document}

\title{Network Effect Decomposition for Causal Effect Estimation under Network Interference}

\author{Zhirong Huang, Wenzhen Zhang, Guangquan Lu, Boyan Chen, Debo Cheng, and Shichao Zhang~\IEEEmembership{Senior Member,~IEEE}
\thanks{Zhirong Huang and Wenzhen Zhang contributed equally to this work.}
\thanks{\textit{Corresponding author: Debo Cheng.}}
\thanks{Zhirong Huang, Wenzhen Zhang, Guangquan Lu, Boyan Chen, and Shichao Zhang are with the School of Computer Science and Engineering, Guangxi Normal University; the Guangxi Key Lab of MIMS; and the MOE Key Lab of EBIT, Guilin, Guangxi 541004, China (e-mail: \href{mailto:huangzhr@foxmail.com}{huangzhr@foxmail.com}, \href{mailto:jianjiu17@outlook.com}{jianjiu17@outlook.com}, \href{mailto:lugq@mailbox.gxnu.edu.cn}{lugq@mailbox.gxnu.edu.cn}, 
\href{mailto:chenboyan@stu.gxnu.edu.cn}{chenboyan@stu.gxnu.edu.cn}, \href{mailto:zhangsc@mailbox.gxnu.edu.cn}{zhangsc@mailbox.gxnu.edu.cn}).}%
\thanks{Debo Cheng is with the School of Computer Science and Technology, Hainan University, Haikou, Hainan 570228, China (e-mail: \href{mailto:chengd@hainanu.edu.cn}{chengd@hainanu.edu.cn}).}}



\maketitle

\begin{abstract}
Estimating causal effects from observational network data requires accounting for both individual treatment and neighborhood exposure. Existing outcome models commonly incorporate these intervention variables into a joint predictor and estimate causal effects through prediction contrasts, leaving the corresponding response components implicit. We propose the \underline{N}etwork \underline{E}ffect \underline{D}ecomposition \underline{Net}work (NEDNet), which parameterizes the conditional outcome response as a query-invariant pre-treatment baseline plus separate treatment- and exposure-response pathways. Path-specific neighborhood contextualization supplies local pre-treatment information to the two pathways. We characterize the fitted response through a unique reference-normalized decomposition, which is equivalent to a zero mixed treatment--exposure contrast. Experiments on semi-synthetic benchmarks derived from BlogCatalog and Flickr show that NEDNet achieves the lowest PEHE among the compared methods for individual, peer, and total effects across homogeneous and covariate-heterogeneous response settings.
\end{abstract}

\begin{IEEEkeywords}
Causal effect estimation, network interference, exposure mapping, potential outcomes.
\end{IEEEkeywords}

\section{Introduction}
\label{sec:intro}
Estimating causal effects from observational data is fundamental to decision-making in public health~\cite{glass2013causal}, social science~\cite{imbens2024causal}, online platforms~\cite{huang2025learning}, and other domains~\cite{yao2021survey}. Conventional causal inference commonly invokes the Stable Unit Treatment Value Assumption (SUTVA), whose no-interference condition requires that a unit's potential outcome be unaffected by other units' treatment assignments~\cite{pearl2009}. This condition may fail in interconnected systems. In social networks~\cite{aronow2017estimating}, spatial communities~\cite{sobel2006randomized}, and economic networks~\cite{forastiere2024causal}, an intervention can affect not only the treated unit but also its neighbors through network connections. An individual's outcome may therefore depend on both its own treatment and the treatments assigned to its neighbors, complicating the characterization of causal effects.

To characterize network-induced interference, an exposure mapping summarizes neighboring treatment assignments into a neighborhood-level variable, allowing potential outcomes to be expressed as functions of individual treatment and neighborhood exposure~\cite{forastiere2021identification,jiang2022estimating}. Under this formulation, the individual effect (IE), peer effect (PE), and total effect (TE) quantify outcome changes associated with individual treatment, neighborhood exposure, or their joint variation~\cite{jiang2022estimating,chen2024doubly}. Estimating these effects requires comparing potential outcomes across alternative treatment--exposure conditions, rather than merely predicting outcomes under observed assignments. This is challenging in observational networks because treatment and exposure assignments may depend on pre-treatment covariates and network structure, whereas outcomes are observed only under realized treatment--exposure combinations. Accurate factual outcome prediction alone therefore does not guarantee accurate estimation of the counterfactual contrasts that define these causal effects.

\begin{figure}[t]
\centering
\includegraphics[width=0.4\linewidth]{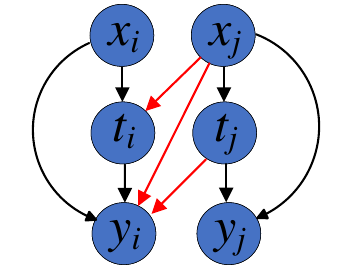}
\caption{Illustrative causal graph for two connected units $i$ and $j$ under network interference. Here, $x$, $t$, and $y$ denote pre-treatment covariates, treatment, and outcome, respectively. Black arrows denote within-unit causal relationships, whereas red arrows denote spillover effects from one unit's treatment to the other unit's outcome.}
\label{fig:causal_graph}
\end{figure}

Learning-based methods for causal effect estimation under network interference can be grouped according to how they incorporate network information into causal modeling. One line of work uses graph-based representations to capture neighborhood dependencies and interference-related context. For example, SPNet combines graph-convolutional representations with an environment summary variable and masked attention to characterize network interference~\cite{huang2023modeling}. Another line incorporates individual treatment and neighborhood exposure into causal estimation, often alongside balancing or robust estimation techniques, as in NetEst, RRNet, and TNet~\cite{jiang2022estimating,cai2023generalization,chen2024doubly}. Related studies further investigate learned exposure mappings and peer-effect decomposition~\cite{adhikari2026learning,du2025telling}. Collectively, these strategies enrich network representations and provide complementary tools for estimating causal effects under network interference.

An important modeling issue is how to explicitly represent the outcome responses associated with individual treatment and neighborhood exposure as distinct components of a conditional outcome model. A joint predictor can incorporate both variables and estimate IE and PE through intervention-based prediction contrasts, but it does not explicitly parameterize the corresponding response components. For example, NetEst incorporates individual treatment and neighborhood exposure into a joint outcome estimator and derives their effects through prediction contrasts~\cite{jiang2022estimating}. gDIS addresses a different aspect of peer-effect analysis by decomposing peer effects into mediation-based direct and indirect pathways; its decomposition concerns causal pathways rather than separately parameterized outcome responses to the two intervention dimensions~\cite{du2025telling}.

To address this issue, we propose the \underline{N}etwork \underline{E}ffect \underline{D}ecomposition \underline{Net}work (NEDNet) for causal effect estimation under network interference. NEDNet represents the conditional outcome response through a query-invariant pre-treatment baseline and separate pathways for individual-treatment and neighborhood-exposure responses. Both pathways are conditioned on compact pre-treatment representations, while path-specific neighborhood contextualization supplies local pre-treatment context. This structured predictor supports potential-outcome estimation under alternative treatment--exposure queries. The main contributions are summarized as follows:
\begin{itemize}
\item A novel response-decomposition framework, termed the Network Effect Decomposition Network (NEDNet), is proposed for conditional outcome modeling under network interference, explicitly separating the pre-treatment baseline from individual-treatment and neighborhood-exposure responses.
\item The separability of NEDNet's conditional outcome function is theoretically characterized under a fixed pre-treatment context, establishing that a unique reference-normalized additive decomposition exists if and only if the mixed treatment--exposure contrast vanishes.
\item The component-wise structure of NEDNet's plug-in estimators for IE, PE, and TE is theoretically characterized, establishing the respective roles of treatment- and exposure-response representations in the estimated effects.
\end{itemize}

\section{Related Work}
\label{sec:related_work}

\subsection{Causal Effect Estimation from Observational Data}

Causal effect estimation from observational data addresses non-random treatment assignment by adjusting for observed pre-treatment covariates~\cite{hirano2003efficient,yao2021survey}. Classical approaches include outcome regression, propensity score methods, inverse probability weighting (IPW), and doubly robust estimators~\cite{rubin1974,rosenbaum1983,robins2000marginal,bang2005doubly}. Their identification typically relies on conditions such as consistency, conditional exchangeability, and positivity, together with the Stable Unit Treatment Value Assumption (SUTVA) in settings without interference.

Recent neural estimators employ representation learning and flexible response functions to model potential outcomes from high-dimensional covariates~\cite{cheng2024disentangled,cheng2023disentangled,cai2024granger,cheng2022toward}. TARNet learns a shared representation with treatment-specific outcome heads, while CFR further balances treated and control representations through an integral probability metric (IPM)~\cite{shalit2017estimating}. Despite their effectiveness for individual treatment-effect estimation, these methods operate under the no-interference setting and do not explicitly account for outcome variations induced by neighboring treatments.

\subsection{Causal Effect Estimation under Network Interference}

Classical studies of causal inference under interference extend the potential-outcome framework to account for treatment spillovers between connected units. Hudgens and Halloran~\cite{hudgens2008toward} formalize direct, indirect, total, and overall causal effects under interference, while Tchetgen Tchetgen and VanderWeele~\cite{tchetgen2012causal} further develop the corresponding identification and estimation framework. Aronow and Samii~\cite{aronow2017estimating} establish a general randomization-based framework for causal effect estimation under interference. For observational networks, Forastiere et al.~\cite{forastiere2021identification,forastiere2022estimating} develop covariate-adjustment and generalized propensity-score methods for estimating treatment and interference effects. 

Some learning-based methods exploit network representations to account for latent confounding and structural imbalance in observational networks~\cite{fan2022debiased,veitch2019using,guo2020learning,chu2021graph}. These approaches primarily focus on confounding adjustment, whereas subsequent studies further investigate the explicit modeling of network interference.

Learning-based studies have incorporated network information into interference-aware causal estimation through several modeling strategies. Graph-based approaches use network structure to represent dependencies among connected units. Ma and Tresp employ graph neural networks for effect estimation under network interference and further study intervention-policy improvement~\cite{ma2021causal}. SPNet combines graph-convolutional representations with an environment summary variable and masked attention to model interference, while CauGramer uses higher-order graph transformers and cross-attention to address unknown interference structures~\cite{huang2023modeling,wu2025causal}. EgoNetGNN takes an exposure-mapping perspective and learns mapping functions for heterogeneous peer effects~\cite{adhikari2026learning}.

Other methods incorporate treatment and neighborhood exposure into causal estimation together with balancing or robust estimation strategies. NetEst combines graph representation learning with adversarial balancing, RRNet integrates joint propensity-score reweighting with IPM-based representation learning, and TNet incorporates targeted learning to construct a doubly robust estimator under network interference~\cite{jiang2022estimating,cai2023generalization,chen2024doubly}. Separately, gDIS decomposes peer effects into mediation-based direct and indirect pathways~\cite{du2025telling}.

These studies advance network representation, interference modeling, exposure mapping, and estimation strategies. NEDNet focuses on the parameterization of the conditional outcome response, representing the pre-treatment baseline and the responses associated with individual treatment and neighborhood exposure as distinct components.

\section{Problem Formulation and Identification}
\label{sec:problem}


\subsection{Networked Observational Data}

Let $\mathcal{G}=(\mathcal{V},\mathcal{E})$ be an undirected network with $N=|\mathcal{V}|$ units and adjacency matrix $A\in\{0,1\}^{N\times N}$, where $A_{ij}=1$ indicates an edge between units $i$ and $j$, $A_{ii}=0$, and $A=A^\top$. For each unit $i\in\mathcal{V}$, we observe a pre-treatment covariate vector $X_i\in\mathbb{R}^{d}$, a binary treatment $T_i\in\{0,1\}$, and a continuous outcome $Y_i\in\mathbb{R}$. Collectively, the observed covariates, treatments, and outcomes are represented by $X=[X_1,\ldots,X_N]^\top$, ${T}=[T_1,\ldots,T_N]^\top$, and ${Y}=[Y_1,\ldots,Y_N]^\top$.

Let $\mathcal{N}_i=\{j\in\mathcal{V}:A_{ij}=1\}$ denote the first-order neighborhood of unit $i$, and let $d_i=|\mathcal{N}_i|$ denote its degree. We write ${T}_{\mathcal{N}_i}=(T_j)_{j\in\mathcal{N}_i}$ and ${X}_{\mathcal{N}_i}=(X_j)_{j\in\mathcal{N}_i}$ for the treatments and covariates of neighboring units, respectively. Throughout this work, the network topology and node covariates are assumed to be fixed prior to treatment assignment, and we consider networks without isolated units so that $d_i>0$ for every unit.

Under network interference, the outcome of a unit may depend on both its own treatment and the treatments assigned to neighboring units. Following the exposure-mapping framework \cite{forastiere2021identification,jiang2022estimating},
we summarize neighboring treatments through the exposure mapping: 
\begin{equation}
Z_i
=
g_i({T}_{\mathcal{N}_i})
=
\frac{1}{d_i}
\sum_{j\in\mathcal{N}_i}T_j,
\label{eq:neighborhood_exposure}
\end{equation}
where $g_i(\cdot)$ denotes the exposure mapping and $Z_i\in[0,1]$ represents the proportion of treated neighbors of unit $i$. The corresponding exposure vector is denoted by ${Z}=[Z_1,\ldots,Z_N]^\top$.


\subsection{Causal Estimands and Problem Definition}

\paragraph{Causal Estimands.}Let $Y_i(t,t_{\mathcal{N}_i})$ denote the potential outcome of unit $i$ under individual treatment $t$ and neighboring treatment assignment $t_{\mathcal{N}_i}$. Under the exposure-mapping assumption, neighboring treatment assignments that induce the same exposure value are outcome-equivalent, allowing us to write $Y_i(t,z)$ for $t\in\{0,1\}$ and $z\in\mathcal{Z}_i$, where $\mathcal{Z}_i=\{k/d_i:k=0,\ldots,d_i\}$ denotes the set of feasible exposure values for a unit with $d_i>0$. Let $C_i=(X_i,X_{\mathcal{N}_i})$ denote the pre-treatment covariates of unit $i$ and its first-order neighbors. The corresponding conditional mean potential outcome is defined as
\begin{equation}
\mu_i(t,z;C_i)
=
\mathbb{E}\!\left[
Y_i(t,z)\mid C_i
\right].
\label{eq:conditional_response}
\end{equation}

For notational simplicity, we suppress $C_i$ in the subsequent expressions, with the understanding that all potential-outcome comparisons for unit $i$ are evaluated under the same fixed pre-treatment context.

Using this conditional response function, we define the IE, PE, and TE as
\begin{align}
\tau_i^{\mathrm{I}}(z)
&=
\mu_i(1,z)-\mu_i(0,z),
\label{eq:individual_effect}\\
\tau_i^{\mathrm{P}}(t;z_a,z_b)
&=
\mu_i(t,z_a)-\mu_i(t,z_b),
\label{eq:peer_effect}\\
\tau_i^{\mathrm{T}}(t_a,z_a;t_b,z_b)
&=
\mu_i(t_a,z_a)-\mu_i(t_b,z_b),
\label{eq:total_effect}
\end{align}
where $t,t_a,t_b\in\{0,1\}$ and $z,z_a,z_b\in\mathcal{Z}_i$. These are unit-level, context-specific contrasts. The IE changes individual treatment while holding neighborhood exposure fixed, whereas the PE changes neighborhood exposure while holding individual treatment fixed. These unit-level contrasts are related to the direct and spillover (or indirect) effects studied in classical interference frameworks~\cite{hudgens2008toward,tchetgen2012causal}. The TE compares two joint treatment--exposure conditions. Population-level summaries can be obtained by averaging the corresponding unit-level contrasts over a specified target population.

\paragraph{Problem Definition.}
Given the observed networked dataset $\mathcal{D}=(A,X,T,Z,Y)$, our goal is to estimate the conditional mean potential outcome $\mu_i(t,z;C_i)$ for each unit under alternative individual treatment and neighborhood exposure conditions, and subsequently derive the corresponding causal effects. Specifically, we seek to learn a response function that approximates
\begin{equation}
\widehat{\mu}_{\theta,i}(t,z;C_i)
\approx
\mu_i(t,z;C_i),
\quad
t\in\{0,1\},\ z\in\mathcal{Z}_i,
\label{eq:response_estimation_target}
\end{equation}
subject to the identification conditions specified below.

Throughout the estimation process, the pre-treatment network structure and covariates are held fixed for each target unit. Accordingly, a counterfactual query consists of an individual treatment and a neighborhood exposure pair $(t,z)$, and neighboring treatments affect the potential outcome only through the specified exposure mapping. The resulting response function therefore characterizes how the outcome of a target unit changes under alternative intervention conditions within its fixed pre-treatment network context.

\subsection{Identification Conditions}

Following established frameworks for causal inference under network interference~\cite{hudgens2008toward,tchetgen2012causal,jiang2022estimating,cai2023generalization,chen2024doubly}, we adopt four identification assumptions: Network Consistency, Neighborhood Interference, Network Overlap, and Network Unconfoundedness. Here, $C_i$ retains the definition given above, with the fixed network topology treated as part of the pre-treatment context. These assumptions specify the conditions under which the target causal contrasts are identifiable from observational data.

\begin{assumption}[Network Consistency]
\label{assump:network_consistency}
For each unit $i$, the observed outcome equals the potential outcome under its observed individual treatment and neighborhood exposure, i.e., $Y_i=Y_i(T_i,Z_i)$.
\end{assumption}

\begin{assumption}[Neighborhood Interference]
\label{assump:neighborhood_interference}
The potential outcome of unit $i$ depends on neighboring treatment assignments only through the exposure mapping $g_i$ applied to its first-order neighborhood. Specifically, for any two neighboring treatment assignments $t_{\mathcal{N}_i}$ and $t'_{\mathcal{N}_i}$ satisfying $g_i(t_{\mathcal{N}_i})=g_i(t'_{\mathcal{N}_i})$, we have $Y_i(t,t_{\mathcal{N}_i})=Y_i(t,t'_{\mathcal{N}_i})$.
\end{assumption}

\begin{assumption}[Network Overlap]
\label{assump:network_overlap}
For each treatment--exposure pair $(t,z)$ required by a target causal contrast, where $t\in\{0,1\}$ and $z\in\mathcal{Z}_i$, $\Pr(T_i=t,Z_i=z\mid C_i)>0$ almost surely in the target population.
\end{assumption}

\begin{assumption}[Network Unconfoundedness]
\label{assump:network_unconfoundedness}
For each admissible treatment--exposure pair $(t,z)$, $Y_i(t,z)\perp\!\!\!\perp(T_i,Z_i)\mid C_i$.
\end{assumption}

Under these assumptions, the conditional mean potential outcome satisfies
\begin{equation}
\begin{aligned}
\mu_i(t,z;C_i) &= \mathbb{E}\!\left[Y_i(t,z)\mid C_i\right] \\
&= \mathbb{E}\!\left[Y_i(t,z)\mid T_i=t,Z_i=z,C_i\right] \\
&= \mathbb{E}\!\left[Y_i\mid T_i=t,Z_i=z,C_i\right].
\end{aligned}
\label{eq:identified_response}
\end{equation}

The second equality follows from Network Unconfoundedness, and the third follows from Network Consistency. Consequently, the IE, PE, and TE in Eqs.~\eqref{eq:individual_effect}--\eqref{eq:total_effect} are identified for contrasts whose treatment--exposure pairs satisfy Network Overlap. The exposure mapping specifies which neighboring assignments are regarded as equivalent, whereas the overlap condition concerns whether a particular pair is supported by the observational assignment distribution. Thus, an exposure level can be feasible under the network structure without having positive observational probability.

\section{The Proposed NEDNet}
\label{sec:method}
\begin{figure*}[t]
\centering
\includegraphics[width=0.9\linewidth]{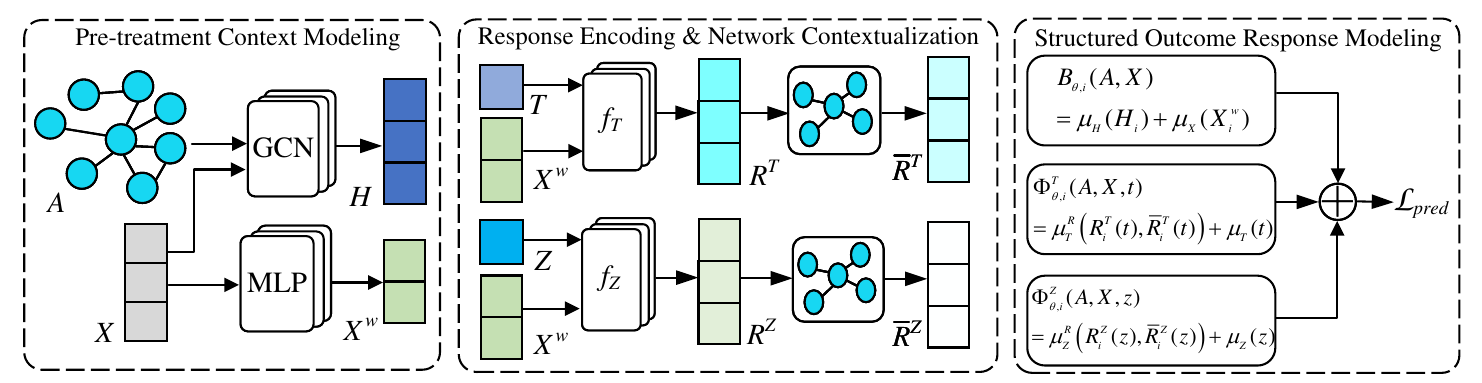}
\caption{Overview of the proposed NEDNet framework. Given the network structure $A$, pre-treatment covariates $X$, individual treatment $T$, and neighborhood exposure $Z$, NEDNet first learns a graph-dependent context representation $H$ and a compact covariate representation $X^{w}$. Treatment and exposure are then processed by separate response encoders and contextualized through separate graph-aggregation operations to obtain $(R^{T},\bar{R}^{T})$ and $(R^{Z},\bar{R}^{Z})$. Finally, the pre-treatment baseline and two intervention-specific response components are integrated to predict factual or queried potential outcomes, from which the IE, PE, and TE are estimated.}
\label{fig:model}
\end{figure*}

As illustrated in Figure~\ref{fig:model}, NEDNet models the conditional outcome response through a pre-treatment baseline component and two intervention-specific response pathways. The treatment and exposure variables are encoded separately to obtain their respective response representations. Each pathway is further contextualized using the target unit's graph representation and neighboring pre-treatment covariates. The resulting components are combined to predict outcomes under alternative treatment--exposure conditions, enabling the estimation of IE, PE, and TE through their corresponding response contrasts.

\subsection{Pre-treatment Context Modeling}
Outcome differences across networked units are not determined solely by treatment and neighborhood exposure. Units may exhibit different baseline outcomes because of their covariates, neighborhood compositions, and structural positions. To represent this query-invariant pre-treatment variation separately from intervention-dependent variation, NEDNet models the graph-dependent baseline and the covariate conditioning information through two representations:
\begin{align}
H_i &= \left[h_{\phi}(A,X)\right]_i, \label{eq:ned_graph_context}\\
X_i^{w} &= \tanh\left(W_xX_i+b_x\right), \label{eq:ned_compact_covariate}
\end{align}
where $h_{\phi}$ is a graph convolutional network (GCN)-based context encoder and $X_i^{w}\in\mathbb{R}^{q}$ is a learned compact covariate representation with $q<d$.

The two representations serve different purposes. The graph-context representation $H_i$ captures topology- and covariate-dependent baseline variation and remains unchanged across alternative treatment--exposure queries. In contrast, $X_i^w$ provides a compact node-specific covariate representation for conditioning both intervention-response pathways. This separation allows graph-dependent context to be modeled in the baseline while retaining explicit covariate conditioning in the response pathways.

\subsection{Treatment and Exposure Response Encoding}

Individual treatment and neighborhood exposure define distinct intervention dimensions in the causal contrasts of Eqs.~\eqref{eq:individual_effect}--\eqref{eq:total_effect}. To retain this distinction during response learning, NEDNet uses two independent encoders conditioned on the compact pre-treatment covariate representation $X_i^{w}$. For a target query $(t,z)$, the local response representations are defined as
\begin{align}
R_i^{T}(t) &= f_T\!\left(t\parallel X_i^{w}\parallel tX_i^{w}\right), \label{eq:ned_treatment_encoder}\\
R_i^{Z}(z) &= f_Z\!\left(z\parallel X_i^{w}\right), \label{eq:ned_exposure_encoder}
\end{align}

Here, $f_T$ and $f_Z$ are path-specific multilayer perceptrons, and $\parallel$ denotes concatenation. The two encoders independently transform individual treatment and neighborhood exposure into their respective response representations while incorporating the same compact pre-treatment covariate information.

The explicit interaction term $tX_i^w$ enables treatment-conditioned feature modulation, allowing the treatment pathway to capture variations in covariate contributions across treatment conditions. Meanwhile, the nonlinear exposure encoder captures exposure-dependent variations in the covariate representation. This path-specific encoding provides distinct representations for modeling individual-treatment and neighborhood-exposure responses.

\subsection{Path-specific Neighborhood Contextualization}

The separate encoders produce intervention-dependent representations conditioned on the target unit's pre-treatment covariates. NEDNet further incorporates neighboring pre-treatment characteristics through path-specific neighborhood contextualization. Specifically, for each target unit $i$, the treatment and exposure pathways aggregate messages constructed from the corresponding response representation, the target unit's graph context, and each neighbor's compact pre-treatment covariate representation:
\begin{align}
\bar{R}_i^{T}(t) &= \frac{1}{\sqrt{d_i}}\sum_{j\in\mathcal{N}_i}\psi_T\!\left(R_i^{T}(t),H_i,X_j^{w}\right), \label{eq:ned_treatment_aggregation}\\
\bar{R}_i^{Z}(z) &= \frac{1}{\sqrt{d_i}}\sum_{j\in\mathcal{N}_i}\psi_Z\!\left(R_i^{Z}(z),H_i,X_j^{w}\right). \label{eq:ned_exposure_aggregation}
\end{align}
Here, $\psi_T$ and $\psi_Z$ are learnable path-specific message functions. Each message combines the target unit's intervention-specific representation and graph context $H_i$ with a neighbor's pre-treatment representation $X_j^w$. The graph context $H_i$ is invariant to the intervention query, whereas $X_j^w$ preserves neighbor-specific covariate information. Separate message functions allow the two pathways to process this context according to their respective intervention representations.

The aggregated representations $\bar{R}_i^T(t)$ and $\bar{R}_i^Z(z)$ supply neighborhood context to their respective response pathways. Individual treatment and neighborhood exposure enter through their corresponding target-unit encoders, while neighboring pre-treatment covariates provide contextual information during aggregation. Neighboring treatment assignments are not supplied separately to the message functions; their influence is represented through the exposure mapping $z$. Each aggregated representation is combined with its local response representation to form the corresponding component in Eqs.~\eqref{eq:ned_treatment_component} and~\eqref{eq:ned_exposure_component}.

\subsection{Structured Outcome Response Modeling}

NEDNet combines the pre-treatment baseline with two intervention-specific response components to construct the conditional outcome predictor. The baseline represents query-invariant outcome variation associated with pre-treatment characteristics and network context, whereas the two response components characterize outcome variations associated with individual treatment and neighborhood exposure, respectively. Specifically, the fitted response is parameterized as
\begin{equation}
\begin{aligned}
\widehat{\mu}_{\theta,i}(t,z;C_i)
&=F_{\theta,i}(A,X;t,z)\\
&=B_{\theta,i}(A,X)+\Phi_{\theta,i}^{T}(t)+\Phi_{\theta,i}^{Z}(z),
\end{aligned}
\label{eq:ned_outcome_decomposition}
\end{equation}
where the three components are defined as
\begin{align}
B_{\theta,i}(A,X) &= \mu_H(H_i)+\mu_X(X_i^{w}), \label{eq:ned_baseline_component}\\
\Phi_{\theta,i}^{T}(t) &= \mu_T^{R}\!\left(R_i^{T}(t),\bar{R}_i^{T}(t)\right)+\mu_T(t), \label{eq:ned_treatment_component}\\
\Phi_{\theta,i}^{Z}(z) &= \mu_Z^{R}\!\left(R_i^{Z}(z),\bar{R}_i^{Z}(z)\right)+\mu_Z(z). \label{eq:ned_exposure_component}
\end{align}

Here, $\mu_H$, $\mu_X$, $\mu_T^{R}$, $\mu_T$, $\mu_Z^{R}$, and $\mu_Z$ are learnable scalar-valued functions. The baseline component combines graph-dependent context with compact pre-treatment covariate information. Each intervention-specific component combines its local response representation and aggregated neighborhood context through a response head, together with a direct readout of the corresponding intervention variable. This organization aligns the model components with the intervention dimensions varied in the IE and PE contrasts. Thus, $\Phi_{\theta,i}^{T}(t)$ and $\Phi_{\theta,i}^{Z}(z)$ are distinct model components for representing outcome variations associated with individual treatment and neighborhood exposure, respectively.

The two response components are defined with respect to their corresponding intervention variables while conditioning on the fixed pre-treatment context. This parameterization determines how NEDNet organizes the fitted outcome response, whereas IE, PE, and TE remain defined by the potential-outcome contrasts in Eqs.~\eqref{eq:individual_effect}--\eqref{eq:total_effect}. The relationship between this response representation and the resulting intervention contrasts is formally characterized in Proposition~\ref{prop:normalized_uniqueness}.

\paragraph{Reference-normalized Response.}
The baseline and response components can be reparameterized through constant shifts without changing the fitted outcome. To obtain a unique representation of these components, we introduce a reference query $(t_0,z_0)$ and normalize the treatment and exposure responses with respect to this query. Throughout this work, we set $(t_0,z_0)=(0,0)$, which is feasible
for every unit because $d_i>0$ and $0\in\mathcal{Z}_i$. Specifically,
\begin{align}
\beta_{\theta,i} &= B_{\theta,i}(A,X)+\Phi_{\theta,i}^{T}(t_0)+\Phi_{\theta,i}^{Z}(z_0), \label{eq:ned_beta}\\
\rho_{\theta,i}^{T}(t) &= \Phi_{\theta,i}^{T}(t)-\Phi_{\theta,i}^{T}(t_0), \notag\\
\rho_{\theta,i}^{Z}(z) &= \Phi_{\theta,i}^{Z}(z)-\Phi_{\theta,i}^{Z}(z_0). \label{eq:ned_rho}
\end{align}
Then the fitted response admits the equivalent representation
\begin{equation}
\begin{aligned}
\widehat{\mu}_{\theta,i}(t,z;C_i)
&=\beta_{\theta,i}+\rho_{\theta,i}^{T}(t)+\rho_{\theta,i}^{Z}(z),\\
\rho_{\theta,i}^{T}(t_0)&=\rho_{\theta,i}^{Z}(z_0)=0.
\end{aligned}
\label{eq:ned_normalized_response}
\end{equation}

Here, $\beta_{\theta,i}$ is the fitted outcome at the reference query, while $\rho_{\theta,i}^{T}(t)$ and $\rho_{\theta,i}^{Z}(z)$ represent the corresponding changes relative to that reference. This normalization is performed on the trained response functions at query time and does not introduce an additional training constraint. Proposition~\ref{prop:normalized_uniqueness} establishes the uniqueness of this reference-normalized representation.

\subsection{Factual Learning Objective}

NEDNet is trained end-to-end using observed factual outcomes. For each training unit $i$, the model evaluates the conditional outcome predictor at its observed treatment--exposure pair $(T_i,Z_i)$, yielding $\widehat{Y}_i=\widehat{\mu}_{\theta,i}(T_i,Z_i;C_i)$. The training objective minimizes the mean squared factual prediction error over the training-node set $\mathcal{V}_{\mathrm{tr}}$:
\begin{equation}
\mathcal{L}_{\mathrm{pred}}
=
\frac{1}{|\mathcal{V}_{\mathrm{tr}}|}
\sum_{i\in\mathcal{V}_{\mathrm{tr}}}
\left(Y_i-\widehat{\mu}_{\theta,i}(T_i,Z_i;C_i)\right)^2.
\label{eq:ned_prediction_loss}
\end{equation}

The objective jointly optimizes the baseline and the two intervention-specific response pathways through factual outcome supervision. After training, the learned response function is evaluated under alternative treatment--exposure queries to obtain counterfactual outcome predictions and estimate the corresponding causal effects. The training procedure is summarized in Algorithm~\ref{alg:nednet}.

\begin{algorithm}[h]
\caption{Training Procedure of NEDNet}
\label{alg:nednet}
\begin{algorithmic}[1]
\Require Adjacency matrix $A$, covariates $X$, treatments $T$,
outcomes $Y$, training nodes $\mathcal{V}_{\mathrm{tr}}$
\Ensure Trained parameters $\theta$

\State $Z\gets\Call{ExposureMapping}{A,T}$
\State Initialize model parameters $\theta$

\For{each training epoch}
    \State $H\gets\Call{ContextEncoder}{A,X}$
    \State $X^{w}\gets\Call{CovariateEncoder}{X}$

    \State $R^{T}\gets\Call{TreatmentEncoder}{T,X^{w}}$
    \State $R^{Z}\gets\Call{ExposureEncoder}{Z,X^{w}}$

    \State $\bar{R}^{T}\gets
    \Call{TreatmentAggregation}{A,R^{T},H,X^{w}}$

    \State $\bar{R}^{Z}\gets
    \Call{ExposureAggregation}{A,R^{Z},H,X^{w}}$

    \State $\mathbf{B}\gets
    \Call{BaselineHead}{H,X^{w}}$

    \State $\boldsymbol{\Phi}^{T}\gets
    \Call{TreatmentHead}{R^{T},\bar{R}^{T},T}$

    \State $\boldsymbol{\Phi}^{Z}\gets
    \Call{ExposureHead}{R^{Z},\bar{R}^{Z},Z}$

    \State $\widehat{\mathbf{Y}}\gets
    \mathbf{B}+\boldsymbol{\Phi}^{T}+\boldsymbol{\Phi}^{Z}$

    \State $\mathcal{L}_{\mathrm{pred}}\gets
    \dfrac{1}{|\mathcal{V}_{\mathrm{tr}}|}
    \sum_{i\in\mathcal{V}_{\mathrm{tr}}}
    (Y_i-\widehat{Y}_i)^2$

    \State Update $\theta$ by minimizing $\mathcal{L}_{\mathrm{pred}}$
\EndFor

\State \Return $\theta$
\end{algorithmic}
\end{algorithm}

\subsection{Potential Outcome Prediction and Effect Estimation}

After training, NEDNet estimates potential outcomes by evaluating the learned response function under alternative treatment--exposure queries. Given a target unit $i$ and a query $(t,z)$, the pre-treatment network context remains fixed, and the predicted potential outcome is expressed using the reference-normalized response representation:
\begin{equation}
\begin{aligned}
\widehat{\mu}_{\theta,i}(t,z;C_i)
&=\left[F_{\theta}(A,X;t,z)\right]_i\\
&=\beta_{\theta,i}+\rho_{\theta,i}^{T}(t)+\rho_{\theta,i}^{Z}(z),
\end{aligned}
\label{eq:ned_query_decomposition}
\end{equation}
where $\beta_{\theta,i}$ represents the reference outcome, and $\rho_{\theta,i}^{T}(t)$ and $\rho_{\theta,i}^{Z}(z)$ characterize the treatment- and exposure-specific response variations, respectively.

Following the causal estimands defined in Eqs.~\eqref{eq:individual_effect}--\eqref{eq:total_effect}, NEDNet estimates the individual, peer, and total effects by substituting the predicted potential outcomes into the corresponding causal contrasts:
\begin{align}
\widehat{\tau}_i^{\mathrm{I}}(z)
&=\widehat{\mu}_{\theta,i}(1,z;C_i)
-\widehat{\mu}_{\theta,i}(0,z;C_i),
\label{eq:ned_ie_estimate}\\
\widehat{\tau}_i^{\mathrm{P}}(t;z_a,z_b)
&=\widehat{\mu}_{\theta,i}(t,z_a;C_i)
-\widehat{\mu}_{\theta,i}(t,z_b;C_i),
\label{eq:ned_pe_estimate}\\
\widehat{\tau}_i^{\mathrm{T}}(t_a,z_a;t_b,z_b)
&=\widehat{\mu}_{\theta,i}(t_a,z_a;C_i)
-\widehat{\mu}_{\theta,i}(t_b,z_b;C_i).
\label{eq:ned_te_estimate}
\end{align}

These plug-in estimators directly correspond to the three causal estimands. Their relationship to the intervention-specific response components is established by the following theoretical analysis.

\subsection{Theoretical Properties of the Structured Response}

Because the treatment and exposure pathways depend on their respective intervention variables, NEDNet's fitted response has a zero mixed treatment--exposure contrast under a fixed pre-treatment context. The following proposition characterizes this property and establishes the uniqueness of the reference-normalized components.

\begin{proposition}[Characterization of Separable Responses]
\label{prop:normalized_uniqueness}
Fix a unit $i$ and its pre-treatment context $C_i$. Let
$f_i(t,z)=\widehat{\mu}_{\theta,i}(t,z;C_i)$ denote the fitted response
over $\{0,1\}\times\mathcal{Z}_i$. Let
$(t_0,z_0)\in\{0,1\}\times\mathcal{Z}_i$ be a fixed reference query.
The following statements are equivalent:

(i) The fitted response admits a separable decomposition
\begin{equation}
f_i(t,z)=\beta_{\theta,i}+\rho_{\theta,i}^{T}(t)+\rho_{\theta,i}^{Z}(z),
\label{eq:separable_response}
\end{equation}
where $\rho_{\theta,i}^{T}(t_0)=\rho_{\theta,i}^{Z}(z_0)=0$ for a fixed reference query $(t_0,z_0)$.

(ii) Its mixed treatment--exposure contrast vanishes for all admissible $t_a,t_b,z_a,z_b$:
\begin{equation}
\begin{aligned}
&f_i(t_a,z_a)-f_i(t_b,z_a)\\
&\quad-f_i(t_a,z_b)+f_i(t_b,z_b)=0.
\end{aligned}
\label{eq:zero_mixed_contrast}
\end{equation}

Moreover, when these conditions hold, the reference-normalized components are uniquely determined by
\begin{align}
\beta_{\theta,i}
&=f_i(t_0,z_0),\notag\\
\rho_{\theta,i}^{T}(t)
&=f_i(t,z_0)-f_i(t_0,z_0),\notag\\
\rho_{\theta,i}^{Z}(z)
&=f_i(t_0,z)-f_i(t_0,z_0).
\label{eq:normalized_components}
\end{align}
\end{proposition}

\begin{proof}
Suppose that $f_i(t,z)$ admits the separable decomposition in Eq.~\eqref{eq:separable_response}. Substituting this decomposition into the mixed treatment--exposure contrast in Eq.~\eqref{eq:zero_mixed_contrast}, we obtain
\begin{align*}
&f_i(t_a,z_a)-f_i(t_b,z_a)
-f_i(t_a,z_b)+f_i(t_b,z_b)\\
&=\rho_{\theta,i}^{T}(t_a)-\rho_{\theta,i}^{T}(t_b)-\rho_{\theta,i}^{T}(t_a)+\rho_{\theta,i}^{T}(t_b)\\
&\quad+\rho_{\theta,i}^{Z}(z_a)-\rho_{\theta,i}^{Z}(z_a)-\rho_{\theta,i}^{Z}(z_b)+\rho_{\theta,i}^{Z}(z_b)\\
&=0.
\end{align*}
Thus, (i) implies (ii).

Conversely, suppose that Eq.~\eqref{eq:zero_mixed_contrast} holds for all admissible intervention queries. Setting $t_b=t_0$ and $z_b=z_0$ yields
\begin{equation}
f_i(t,z)=f_i(t,z_0)+f_i(t_0,z)-f_i(t_0,z_0).
\end{equation}
Defining the three components according to Eq.~\eqref{eq:normalized_components} directly gives the separable decomposition in Eq.~\eqref{eq:separable_response}, with the required reference constraints. Hence, (ii) implies (i).

It remains to establish uniqueness. The reference constraints uniquely determine the baseline as $f_i(t_0,z_0)$, the treatment-response component as $f_i(t,z_0)-f_i(t_0,z_0)$, and the exposure-response component as $f_i(t_0,z)-f_i(t_0,z_0)$. Therefore, the reference-normalized decomposition is unique.
\end{proof}

\begin{corollary}[Effect Contrast Separation]
\label{cor:effect_contrast_separation}
For the fitted response of NEDNet, the estimated IE, PE, and TE satisfy
\begin{align}
\widehat{\tau}_i^{\mathrm{I}}(z)
&=\rho_{\theta,i}^{T}(1)-\rho_{\theta,i}^{T}(0),\\
\widehat{\tau}_i^{\mathrm{P}}(t;z_a,z_b)
&=\rho_{\theta,i}^{Z}(z_a)-\rho_{\theta,i}^{Z}(z_b),\\
\widehat{\tau}_i^{\mathrm{T}}(t_a,z_a;t_b,z_b)
&=\rho_{\theta,i}^{T}(t_a)-\rho_{\theta,i}^{T}(t_b)\notag\\
&\quad+\rho_{\theta,i}^{Z}(z_a)-\rho_{\theta,i}^{Z}(z_b).
\end{align}
Accordingly, the plug-in estimate of IE obtained from NEDNet does not vary with the exposure query, while the plug-in estimate of PE does not vary with the individual-treatment query. The plug-in estimate of TE is given by the sum of the corresponding treatment- and exposure-response contrasts.
\end{corollary}

\begin{proof}
By Proposition~\ref{prop:normalized_uniqueness}, the fitted response admits the reference-normalized decomposition
\begin{equation}
f_i(t,z)=\beta_{\theta,i}+\rho_{\theta,i}^{T}(t)+\rho_{\theta,i}^{Z}(z).
\end{equation}
Substituting this expression into the plug-in IE gives
\begin{equation}
\begin{aligned}
\widehat{\tau}_i^{\mathrm{I}}(z)
&=f_i(1,z)-f_i(0,z)\\
&=\rho_{\theta,i}^{T}(1)-\rho_{\theta,i}^{T}(0).
\end{aligned}
\end{equation}
Similarly, the plug-in PE is
\begin{equation}
\begin{aligned}
\widehat{\tau}_i^{\mathrm{P}}(t;z_a,z_b)
&=f_i(t,z_a)-f_i(t,z_b)\\
&=\rho_{\theta,i}^{Z}(z_a)-\rho_{\theta,i}^{Z}(z_b).
\end{aligned}
\end{equation}
Finally, substituting the same decomposition into the plug-in TE yields
\begin{equation}
\begin{aligned}
\widehat{\tau}_i^{\mathrm{T}}(t_a,z_a;t_b,z_b)
&=f_i(t_a,z_a)-f_i(t_b,z_b)\\
&=\rho_{\theta,i}^{T}(t_a)-\rho_{\theta,i}^{T}(t_b)\\
&\quad+\rho_{\theta,i}^{Z}(z_a)-\rho_{\theta,i}^{Z}(z_b).
\end{aligned}
\end{equation}
The stated properties follow directly from these expressions.
\end{proof}

\section{Experiments}
\label{sec:experiments}

We evaluate NEDNet on counterfactual outcome and causal effect estimation, and examine the contributions of its architectural components through ablation studies.

\subsection{Datasets and Semi-synthetic Data Generation}
\label{sec:dgp}

Following the benchmark settings used in NetEst~\cite{jiang2022estimating} and TNet~\cite{chen2024doubly}, we use two real-world social networks, BlogCatalog (BC) and Flickr, to construct semi-synthetic benchmarks. BC contains blogger relationships and keyword-based node features, whereas Flickr contains user connections and interest-tag features. We retain the original network structures $A$ and node covariates $X$, while simulating treatment assignments and potential outcomes to obtain ground-truth counterfactual outcomes and causal effects.

To prevent information exchange between training, validation, and test nodes through graph message passing, each network is divided into vertex-disjoint training, validation, and test subgraphs. All neighborhood-dependent operations, including degree calculations, neighborhood averages, treatment assignment, exposure computation, and graph aggregation, are restricted to the corresponding subgraph. Isolated nodes are removed, resulting in $5{,}196$ nodes for BC and $7{,}301$ nodes for Flickr.

For each independent run, we sample $W_1\sim\mathcal{N}(0,I_d)$ and compute the individual and neighborhood treatment scores:
\begin{equation}
p_i^T=\sigma(W_1^\top X_i),
\qquad
p_{\mathcal{N}_i}^T
=
\frac{1}{d_i}
\sum_{j\in\mathcal{N}_i}p_j^T,
\label{eq:dgp_treatment_scores}
\end{equation}
where $\sigma(\cdot)$ denotes the sigmoid function. The overall treatment score is $s_i^T=p_i^T+p_{\mathcal{N}_i}^T$. Let $\mathcal{V}_s$ denote the graph partition containing node $i$. The binary treatment is assigned according to
\begin{equation}
T_i
=
\mathbb{I}\left(s_i^T>\overline{s}_s^T\right),
\qquad
\overline{s}_s^T
=
\frac{1}{|\mathcal{V}_s|}
\sum_{k\in\mathcal{V}_s}s_k^T.
\label{eq:dgp_treatment_assignment}
\end{equation}
Given the realized treatment assignments, the neighborhood exposure is calculated as
\begin{equation}
Z_i
=
\frac{1}{d_i}
\sum_{j\in\mathcal{N}_i}T_j.
\label{eq:dgp_exposure}
\end{equation}

For potential outcome generation, we independently sample $W_2\sim\mathcal{N}(0,I_d)$ and compute
\begin{equation}
p_i^Y=\sigma(W_2^\top X_i),
\qquad
p_{\mathcal{N}_i}^Y
=
\frac{1}{d_i}
\sum_{j\in\mathcal{N}_i}p_j^Y.
\label{eq:dgp_outcome_scores}
\end{equation}
We consider three outcome-generation settings. The homogeneous setting follows the original semi-synthetic benchmark, while the two heterogeneous settings follow TNet~\cite{chen2024doubly}. Specifically, the potential outcome is defined as

\begin{equation}
Y_i(t,z)=
\begin{cases}
t+z+p_i^Y+0.5p_{\mathcal{N}_i}^Y+\epsilon_i,
& \texttt{homo},\\\\
\begin{aligned}
&t+z+p_i^Y+0.5p_{\mathcal{N}_i}^Y\\
&\quad+t(p_i^Y+0.5p_{\mathcal{N}_i}^Y)+\epsilon_i,
\end{aligned}
& \texttt{hete},\\\\
\begin{aligned}
&t+z+p_i^Y+0.5p_{\mathcal{N}_i}^Y\\
&\quad+t(p_i^Y+0.5p_{\mathcal{N}_i}^Y)\\
&\quad+z(0.5p_i^Y+p_{\mathcal{N}_i}^Y)+\epsilon_i,
\end{aligned}
& \texttt{hete\_z}.
\end{cases}
\label{eq:dgp_potential_outcome}
\end{equation}

Here, $\epsilon_i$ is a Gaussian disturbance, sampled once for each node and held fixed across all treatment--exposure queries within the same run. The observed factual outcome is therefore $Y_i=Y_i(T_i,Z_i)$. The \texttt{hete} setting introduces heterogeneous individual-treatment effects through node-dependent covariate modulation. The \texttt{hete\_z} setting further incorporates heterogeneous neighborhood-exposure effects through an additional covariate-dependent response term.

We conduct five independent runs, indexed by expID. For each run, $W_1$, $W_2$, and the node-specific disturbances $\{\epsilon_i\}$ are independently resampled. Within each run, the sampled disturbances remain fixed across all treatment--exposure queries. All competing methods use identical graph partitions, treatment assignments, factual outcomes, and ground-truth potential outcomes within each run.

\subsection{Experimental Setup}
\label{sec:experimental_setup}

\paragraph{Baselines.}
We compare NEDNet with seven representative estimators, including exposure-augmented outcome models and methods developed for networked causal inference. \textbf{TARNet+Z} and \textbf{CFR+Z}~\cite{jiang2022estimating} extend TARNet and CFR~\cite{shalit2017estimating} by incorporating neighborhood exposure as an additional outcome-model input. \textbf{NetDeconf+Z}~\cite{jiang2022estimating} similarly augments the Network Deconfounder~\cite{guo2020learning} with an explicit exposure input. \textbf{NetEst}~\cite{jiang2022estimating} combines graph representation learning with balancing objectives for treatment and exposure. \textbf{RRNet}~\cite{cai2023generalization} integrates network representation learning with propensity-score reweighting, whereas \textbf{TNet}~\cite{chen2024doubly} uses targeted learning to construct a doubly robust estimator. We also include \textbf{gDIS}~\cite{du2025telling}, which decomposes peer effects through graph-based mediation analysis.

\begin{figure*}[ht]
\centering
\includegraphics[width=\linewidth]{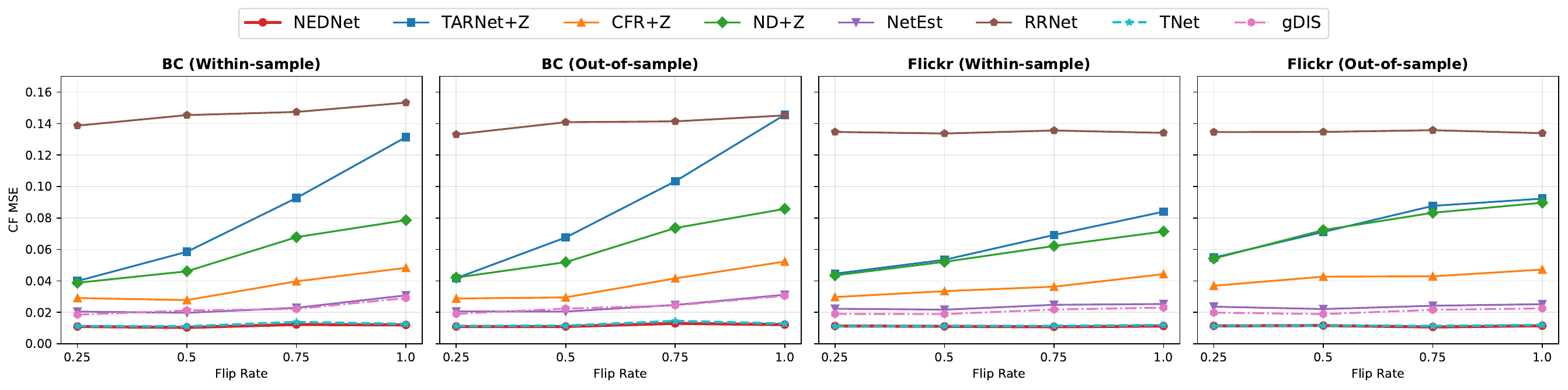}
\caption{Counterfactual outcome estimation performance (CF-MSE) under different treatment Flip Rates on BC and Flickr with the homogeneous outcome-generation setting. From left to right, the panels show within-sample and out-of-sample results on BC, followed by those on Flickr. Lower values indicate better performance.}
\label{fig:cf_fliprate}
\end{figure*}

\begin{table*}[t]
\centering
\caption{Causal effect estimation results on the homogeneous (homo) datasets measured by $\sqrt{\epsilon_{\mathrm{PEHE}}}$. Results are reported as mean $\pm$ standard deviation over five independent runs. Lower values are better; the best and second-best results are highlighted in bold and underlined, respectively. ``Ind.'', ``Peer'', and ``Total'' denote the Individual, Peer, and Total Effects.}
\label{tab:main_effect_results}
\resizebox{\textwidth}{!}{
\begin{tabular}{clllllllll}
\toprule
Datasets &
Effects &
TARNet+Z &
CFR+Z &
NetDeconf+Z &
NetEst &
RRNet &
TNet &
gDIS &
NEDNet \\
\midrule

\multirow{3}{*}{\begin{tabular}[c]{@{}c@{}}BC\\ (Within-sample)\end{tabular}}
& Ind.
& $0.2000_{\pm0.0674}$
& $0.1506_{\pm0.0678}$
& $0.2690_{\pm0.1379}$
& $\underline{0.0782_{\pm0.0249}}$
& $0.0972_{\pm0.0915}$
& $0.0809_{\pm0.0360}$
& $0.1287_{\pm0.0882}$
& $\mathbf{0.0472_{\pm0.0319}}$ \\

& Peer
& $0.4977_{\pm0.0520}$
& $0.3854_{\pm0.0857}$
& $0.4962_{\pm0.1162}$
& $0.0782_{\pm0.0336}$
& $\underline{0.0501_{\pm0.0248}}$
& $0.0823_{\pm0.0337}$
& $0.0922_{\pm0.0474}$
& $\mathbf{0.0251_{\pm0.0098}}$ \\

& Total
& $0.7855_{\pm0.1743}$
& $0.5132_{\pm0.1986}$
& $0.6457_{\pm0.2817}$
& $0.1871_{\pm0.0898}$
& $0.1682_{\pm0.0843}$
& $\underline{0.0860_{\pm0.0464}}$
& $0.2428_{\pm0.1805}$
& $\mathbf{0.0539_{\pm0.0292}}$ \\
\midrule

\multirow{3}{*}{\begin{tabular}[c]{@{}c@{}}BC\\ (Out-of-sample)\end{tabular}}
& Ind.
& $0.2058_{\pm0.0686}$
& $0.1521_{\pm0.0683}$
& $0.2635_{\pm0.1365}$
& $\underline{0.0706_{\pm0.0278}}$
& $0.0957_{\pm0.0951}$
& $0.0819_{\pm0.0356}$
& $0.1308_{\pm0.0880}$
& $\mathbf{0.0504_{\pm0.0342}}$ \\

& Peer
& $0.4958_{\pm0.0550}$
& $0.3853_{\pm0.0858}$
& $0.4962_{\pm0.1164}$
& $0.0742_{\pm0.0352}$
& $\underline{0.0501_{\pm0.0250}}$
& $0.0821_{\pm0.0336}$
& $0.0941_{\pm0.0455}$
& $\mathbf{0.0256_{\pm0.0097}}$ \\

& Total
& $0.7802_{\pm0.1745}$
& $0.5124_{\pm0.2004}$
& $0.6342_{\pm0.2842}$
& $0.1786_{\pm0.0874}$
& $0.1667_{\pm0.0800}$
& $\underline{0.0884_{\pm0.0482}}$
& $0.2522_{\pm0.1732}$
& $\mathbf{0.0568_{\pm0.0327}}$ \\
\midrule

\multirow{3}{*}{\begin{tabular}[c]{@{}c@{}}Flickr\\ (Within-sample)\end{tabular}}
& Ind.
& $0.2051_{\pm0.0419}$
& $0.1643_{\pm0.0609}$
& $0.2776_{\pm0.0902}$
& $0.1363_{\pm0.0576}$
& $0.1049_{\pm0.0365}$
& $\underline{0.0395_{\pm0.0227}}$
& $0.0680_{\pm0.0228}$
& $\mathbf{0.0276_{\pm0.0108}}$ \\

& Peer
& $0.4794_{\pm0.0281}$
& $0.3716_{\pm0.0510}$
& $0.4939_{\pm0.0638}$
& $0.0789_{\pm0.0318}$
& $0.0528_{\pm0.0112}$
& $0.0857_{\pm0.0179}$
& $\underline{0.0384_{\pm0.0062}}$
& $\mathbf{0.0196_{\pm0.0048}}$ \\

& Total
& $0.7835_{\pm0.0725}$
& $0.5479_{\pm0.0746}$
& $0.6934_{\pm0.0935}$
& $0.2654_{\pm0.0695}$
& $0.1573_{\pm0.0365}$
& $\underline{0.0551_{\pm0.0209}}$
& $0.1231_{\pm0.0665}$
& $\mathbf{0.0443_{\pm0.0268}}$ \\
\midrule

\multirow{3}{*}{\begin{tabular}[c]{@{}c@{}}Flickr\\ (Out-of-sample)\end{tabular}}
& Ind.
& $0.1793_{\pm0.0351}$
& $0.1647_{\pm0.0611}$
& $0.2585_{\pm0.0892}$
& $0.1235_{\pm0.0508}$
& $0.1050_{\pm0.0360}$
& $\underline{0.0386_{\pm0.0229}}$
& $0.0775_{\pm0.0184}$
& $\mathbf{0.0267_{\pm0.0081}}$ \\

& Peer
& $0.4681_{\pm0.0330}$
& $0.3716_{\pm0.0510}$
& $0.4938_{\pm0.0640}$
& $0.0661_{\pm0.0307}$
& $0.0526_{\pm0.0106}$
& $0.0857_{\pm0.0178}$
& $\underline{0.0487_{\pm0.0103}}$
& $\mathbf{0.0204_{\pm0.0044}}$ \\

& Total
& $0.7113_{\pm0.0573}$
& $0.5475_{\pm0.0757}$
& $0.6703_{\pm0.0959}$
& $0.2350_{\pm0.0574}$
& $0.1563_{\pm0.0355}$
& $\underline{0.0540_{\pm0.0207}}$
& $0.1390_{\pm0.0636}$
& $\mathbf{0.0428_{\pm0.0224}}$ \\
\bottomrule
\end{tabular}}
\end{table*}

\paragraph{Counterfactual Outcome Evaluation.}
Following NetEst~\cite{jiang2022estimating}, RRNet~\cite{cai2023generalization}, and TNet~\cite{chen2024doubly}, we evaluate counterfactual outcome prediction by randomly selecting a proportion of nodes and reversing their factual treatments. Let $\mathcal{V}_{\mathrm{flip}}$ denote the selected node set. The counterfactual treatment assignment is defined as
\begin{equation}
T_i^{\mathrm{cf}}
=
\begin{cases}
1-T_i, & i\in\mathcal{V}_{\mathrm{flip}},\\
T_i, & i\notin\mathcal{V}_{\mathrm{flip}}.
\end{cases}
\label{eq:counterfactual_treatment}
\end{equation}
The ratio $|\mathcal{V}_{\mathrm{flip}}|/|\mathcal{V}|$ is referred to as the \emph{Flip Rate}. Since treatment changes can also alter the exposures of neighboring nodes, the counterfactual neighborhood exposure is recomputed as
\begin{equation}
Z_i^{\mathrm{cf}}
=
\frac{1}{d_i}
\sum_{j\in\mathcal{N}_i}T_j^{\mathrm{cf}}.
\label{eq:counterfactual_exposure}
\end{equation}
The corresponding ground-truth counterfactual outcome is obtained from the known DGP:
\begin{equation}
Y_i^{\mathrm{cf}}
=
Y_i(T_i^{\mathrm{cf}},Z_i^{\mathrm{cf}}).
\label{eq:counterfactual_outcome}
\end{equation}
NEDNet predicts the counterfactual outcome by evaluating its learned response function under the counterfactual treatment--exposure pair:
\begin{equation}
\widehat{Y}_i^{\mathrm{cf}}
=
\widehat{\mu}_{\theta,i}
(T_i^{\mathrm{cf}},Z_i^{\mathrm{cf}};C_i).
\label{eq:counterfactual_prediction}
\end{equation}
The counterfactual prediction error is measured by
\begin{equation}
\mathrm{CF\text{-}MSE}
=
\frac{1}{|\mathcal{V}_{\mathrm{eval}}|}
\sum_{i\in\mathcal{V}_{\mathrm{eval}}}
\left(
\widehat{Y}_i^{\mathrm{cf}}
-
Y_i^{\mathrm{cf}}
\right)^2.
\label{eq:counterfactual_mse}
\end{equation}

\paragraph{Causal Effect Evaluation.}
For causal effect estimation, the ground-truth IE, PE, and TE are calculated from the potential outcomes generated by the DGP, while their estimates are obtained through the plug-in contrasts of the predicted potential outcomes. For each effect type $e\in\{\mathrm{I},\mathrm{P},\mathrm{T}\}$, we report the Precision in Estimation of Heterogeneous Effect (PEHE):
\begin{equation}
\sqrt{\epsilon_{\mathrm{PEHE}}^{\,e}}
=
\sqrt{
\frac{1}{|\mathcal{V}_{\mathrm{eval}}|}
\sum_{i\in\mathcal{V}_{\mathrm{eval}}}
\left(
\widehat{\tau}_i^{e}-\tau_i^{e}
\right)^2
}.
\label{eq:effect_pehe}
\end{equation}
Lower CF-MSE and PEHE values indicate better counterfactual outcome and causal effect estimation, respectively. We report within-sample performance on the training subgraphs and out-of-sample performance on the held-out test subgraphs.

\paragraph{Implementation Details.}
NEDNet is implemented in PyTorch and evaluated on a server equipped with a 64-core Montage Jintide C6326 CPU and eight NVIDIA GeForce RTX 4090 GPUs with 24\,GB of memory each. The same hyperparameter configuration is used for BlogCatalog and Flickr. The graph-context encoder contains one graph convolutional network (GCN) layer followed by a two-layer node-wise multilayer perceptron (MLP), while the dimension of the compact covariate representation is fixed at $2$. The hidden dimensions of the GCN and path-specific MLP encoders are selected from prespecified candidate ranges according to the factual outcome prediction error on the validation set. NEDNet is trained for $500$ epochs using the Adam optimizer with a learning rate of $10^{-3}$ and no weight decay.

For competing methods, we adopt the recommended hyperparameter settings reported in their original papers or released implementations. For TARNet+Z, CFR+Z, and NetDeconf+Z, we retain the original architectures and hyperparameters while incorporating neighborhood exposure as an additional input. All methods are evaluated using identical data realizations and graph partitions.

\begin{table*}[t]
\centering
\caption{Ablation results on BC and Flickr measured by $\sqrt{\epsilon_{\mathrm{PEHE}}}$.}
\label{tab:ablation_results}
\resizebox{\textwidth}{!}{
\begin{tabular}{cl|lllll}
\toprule
Datasets & Effects & NEDNet & w/o T-Rep. & w/o Z-Rep. & Shared encoder & Joint decoder \\
\midrule

\multirow{3}{*}{\begin{tabular}[c]{@{}c@{}}BC\\ (Within-sample)\end{tabular}}
& Ind.
& $\mathbf{0.0472_{\pm0.0319}}$
& $1.2008_{\pm0.0254}$
& $2.0336_{\pm0.0667}$
& $0.0678_{\pm0.0406}$
& $\underline{0.0555_{\pm0.0213}}$ \\

& Peer
& $\mathbf{0.0251_{\pm0.0098}}$
& $0.6734_{\pm0.1077}$
& $0.7929_{\pm0.0286}$
& $\underline{0.0283_{\pm0.0076}}$
& $0.0367_{\pm0.0104}$ \\

& Total
& $\mathbf{0.0539_{\pm0.0292}}$
& $0.7512_{\pm0.0448}$
& $0.8665_{\pm0.0341}$
& $\underline{0.0761_{\pm0.0355}}$
& $0.0870_{\pm0.0400}$ \\
\midrule

\multirow{3}{*}{\begin{tabular}[c]{@{}c@{}}BC\\ (Out-of-sample)\end{tabular}}
& Ind.
& $\mathbf{0.0504_{\pm0.0342}}$
& $1.2019_{\pm0.0279}$
& $1.8051_{\pm0.0629}$
& $0.0689_{\pm0.0417}$
& $\underline{0.0583_{\pm0.0218}}$ \\

& Peer
& $\mathbf{0.0256_{\pm0.0097}}$
& $0.6203_{\pm0.0805}$
& $0.7932_{\pm0.0276}$
& $\underline{0.0294_{\pm0.0083}}$
& $0.0372_{\pm0.0102}$ \\

& Total
& $\mathbf{0.0568_{\pm0.0327}}$
& $0.7359_{\pm0.0884}$
& $0.5933_{\pm0.0173}$
& $\underline{0.0773_{\pm0.0351}}$
& $0.0862_{\pm0.0356}$ \\
\midrule

\multirow{3}{*}{\begin{tabular}[c]{@{}c@{}}Flickr\\ (Within-sample)\end{tabular}}
& Ind.
& $\mathbf{0.0276_{\pm0.0108}}$
& $1.1896_{\pm0.0179}$
& $3.2258_{\pm0.1623}$
& $0.0387_{\pm0.0107}$
& $\underline{0.0347_{\pm0.0069}}$ \\

& Peer
& $\mathbf{0.0196_{\pm0.0048}}$
& $0.9180_{\pm0.2425}$
& $0.7479_{\pm0.0141}$
& $\underline{0.0285_{\pm0.0071}}$
& $0.0303_{\pm0.0070}$ \\

& Total
& $0.0443_{\pm0.0268}$
& $1.1341_{\pm0.4282}$
& $2.1801_{\pm0.1445}$
& $\underline{0.0440_{\pm0.0091}}$
& $\mathbf{0.0429_{\pm0.0050}}$ \\
\midrule

\multirow{3}{*}{\begin{tabular}[c]{@{}c@{}}Flickr\\ (Out-of-sample)\end{tabular}}
& Ind.
& $\mathbf{0.0267_{\pm0.0081}}$
& $1.1906_{\pm0.0179}$
& $2.3737_{\pm0.1170}$
& $0.0386_{\pm0.0118}$
& $\underline{0.0340_{\pm0.0093}}$ \\

& Peer
& $\mathbf{0.0204_{\pm0.0044}}$
& $0.7206_{\pm0.1616}$
& $0.7487_{\pm0.0149}$
& $\underline{0.0294_{\pm0.0066}}$
& $0.0306_{\pm0.0061}$ \\

& Total
& $\mathbf{0.0428_{\pm0.0224}}$
& $0.8939_{\pm0.2097}$
& $1.4853_{\pm0.0913}$
& $\underline{0.0429_{\pm0.0080}}$
& $0.0438_{\pm0.0073}$ \\
\bottomrule
\end{tabular}}
\end{table*}

\subsection{Main Results}
\label{sec:main_results}

\paragraph{Counterfactual Outcome Estimation.}
Figure~\ref{fig:cf_fliprate} reports the counterfactual prediction errors under different treatment Flip Rates. Increasing the Flip Rate changes the treatment assignments of more nodes and consequently alters the neighborhood exposures induced by the network. NEDNet maintains consistently low CF-MSE across the evaluated Flip Rates on both datasets. Moreover, its within-sample and out-of-sample performance remains comparable, indicating that the learned response function generalizes to held-out graph subgraphs under alternative treatment assignments.

\paragraph{Causal Effect Estimation.}
Table~\ref{tab:main_effect_results} reports the estimation errors of IE, PE, and TE on BC and Flickr. NEDNet achieves the lowest PEHE in all twelve dataset--setting--effect combinations. The strongest competing method varies across effect types and datasets, suggesting that the relative performance of existing estimators depends on the particular causal contrast being evaluated. In comparison, NEDNet consistently provides the most accurate estimates across all three effects.

The advantage is particularly evident in the out-of-sample results. Compared with the best-performing baseline in each setting, NEDNet reduces the estimation error by $28.6\%$--$48.9\%$ on BC and $20.7\%$--$58.1\%$ on Flickr. The reductions are especially pronounced for PE, with improvements of $48.9\%$ on BC and $58.1\%$ on Flickr. These results demonstrate the effectiveness of NEDNet in estimating neighborhood-exposure effects while maintaining strong performance in individual and joint intervention effect estimation.

The gaps between within-sample and out-of-sample PEHE remain modest across the three effect types. This pattern suggests that the estimation gains are not confined to the nodes used for model fitting and are largely maintained on held-out graph subgraphs.

\subsection{Ablation Study}
\label{sec:ablation}

We conduct ablation studies on BC and Flickr to examine the contributions of the response representations, path-specific encoding, and decoder design. Four variants are considered. \textit{w/o T-Rep.} and \textit{w/o Z-Rep.} remove the representation-based terms $\mu_T^R$ and $\mu_Z^R$, respectively, while retaining the corresponding direct readouts $\mu_T$ and $\mu_Z$. \textit{Shared encoder} jointly encodes treatment and exposure before graph contextualization, whereas \textit{Joint decoder} replaces the path-specific response decoding with a unified MLP. All variants are evaluated under identical training and evaluation settings.

Table~\ref{tab:ablation_results} reports the ablation results on BC and Flickr. Removing either representation-based response term leads to substantial increases in PEHE across the three effect types and both evaluation settings. These results indicate that the graph-contextualized treatment and exposure representations provide complementary information beyond the corresponding direct readouts, contributing to the estimation of individual, peer, and total effects.

The comparison with the shared encoder further highlights the importance of path-specific encoding. NEDNet consistently achieves lower errors than the shared-encoder variant across all BC settings and the individual and peer effects on Flickr. A similar pattern is observed for the joint decoder, which generally produces higher errors than NEDNet. These results suggest that separately encoding the two intervention dimensions and preserving their respective response pathways are beneficial to causal effect estimation.

On Flickr, the joint decoder and shared encoder obtain slightly lower within-sample errors for the total effect than NEDNet, with PEHE values of $0.0429$ and $0.0440$, respectively, compared with $0.0443$ for NEDNet. However, neither variant achieves lower errors for the individual and peer effects in this setting, and both yield higher total-effect errors in the out-of-sample evaluation. This result indicates that the relative performance of decoder configurations can vary across effect types and evaluation settings, while the path-specific design of NEDNet provides consistently strong performance across both datasets.

\begin{table*}[!ht]
\centering
\caption{Causal effect estimation on the \texttt{hete} datasets measured by $\sqrt{\epsilon_{\mathrm{PEHE}}}$. TNet$^\dagger$ denotes TNet with hyperparameters selected on the validation set for the corresponding heterogeneous setting.}
\label{tab:hetero_hete_effect}
\resizebox{\textwidth}{!}{
\begin{tabular}{clllllllll}
\toprule
Datasets & Effects & TARNet+Z & CFR+Z & NetDeconf+Z & NetEst & RRNet & TNet$^\dagger$ & gDIS & NEDNet \\
\midrule
\multirow{3}{*}{\begin{tabular}[c]{@{}c@{}}BC\\(Within-sample)\end{tabular}}
& Ind. & $0.3438_{\pm0.0876}$ & $0.3240_{\pm0.0703}$ & $0.4856_{\pm0.1378}$ & $0.1571_{\pm0.0532}$ & $0.1108_{\pm0.0446}$ & $\underline{0.0567_{\pm0.0217}}$ & $0.2691_{\pm0.0635}$ & $\mathbf{0.0353_{\pm0.0064}}$ \\
& Peer & $0.5588_{\pm0.0540}$ & $0.5052_{\pm0.0596}$ & $0.6018_{\pm0.0666}$ & $0.0778_{\pm0.0269}$ & $\underline{0.0501_{\pm0.0190}}$ & $0.1037_{\pm0.0324}$ & $0.1398_{\pm0.0672}$ & $\mathbf{0.0229_{\pm0.0099}}$ \\
& Total & $0.8413_{\pm0.1467}$ & $0.6859_{\pm0.1688}$ & $0.7586_{\pm0.1875}$ & $0.2189_{\pm0.0918}$ & $0.2035_{\pm0.0584}$ & $\underline{0.1265_{\pm0.0827}}$ & $0.5089_{\pm0.1609}$ & $\mathbf{0.0636_{\pm0.0316}}$ \\
\midrule
\multirow{3}{*}{\begin{tabular}[c]{@{}c@{}}BC\\(Out-of-sample)\end{tabular}}
& Ind. & $0.3444_{\pm0.0942}$ & $0.3278_{\pm0.0660}$ & $0.4781_{\pm0.1398}$ & $0.1519_{\pm0.0582}$ & $0.1091_{\pm0.0433}$ & $\underline{0.0562_{\pm0.0217}}$ & $0.2724_{\pm0.0657}$ & $\mathbf{0.0361_{\pm0.0069}}$ \\
& Peer & $0.5590_{\pm0.0553}$ & $0.5052_{\pm0.0596}$ & $0.6018_{\pm0.0665}$ & $0.0734_{\pm0.0305}$ & $\underline{0.0501_{\pm0.0201}}$ & $0.1027_{\pm0.0301}$ & $0.1409_{\pm0.0662}$ & $\mathbf{0.0235_{\pm0.0099}}$ \\
& Total & $0.8236_{\pm0.1599}$ & $0.6885_{\pm0.1689}$ & $0.7466_{\pm0.1898}$ & $0.2110_{\pm0.0912}$ & $0.2026_{\pm0.0551}$ & $\underline{0.1281_{\pm0.0845}}$ & $0.5120_{\pm0.1592}$ & $\mathbf{0.0663_{\pm0.0330}}$ \\
\midrule
\multirow{3}{*}{\begin{tabular}[c]{@{}c@{}}Flickr\\(Within-sample)\end{tabular}}
& Ind. & $0.3709_{\pm0.0490}$ & $0.3305_{\pm0.0807}$ & $0.4738_{\pm0.0467}$ & $0.2230_{\pm0.0561}$ & $0.1208_{\pm0.0278}$ & $\underline{0.0754_{\pm0.0282}}$ & $0.2326_{\pm0.0446}$ & $\mathbf{0.0481_{\pm0.0352}}$ \\
& Peer & $0.5287_{\pm0.0361}$ & $0.4451_{\pm0.0574}$ & $0.5657_{\pm0.0262}$ & $0.0637_{\pm0.0165}$ & $\underline{0.0543_{\pm0.0067}}$ & $0.0977_{\pm0.0308}$ & $0.0985_{\pm0.0266}$ & $\mathbf{0.0222_{\pm0.0088}}$ \\
& Total & $0.8385_{\pm0.0924}$ & $0.6236_{\pm0.0960}$ & $0.7545_{\pm0.0750}$ & $0.3105_{\pm0.0740}$ & $0.1942_{\pm0.0316}$ & $\underline{0.1185_{\pm0.0365}}$ & $0.4411_{\pm0.0784}$ & $\mathbf{0.0530_{\pm0.0207}}$ \\
\midrule
\multirow{3}{*}{\begin{tabular}[c]{@{}c@{}}Flickr\\(Out-of-sample)\end{tabular}}
& Ind. & $0.3343_{\pm0.0406}$ & $0.3277_{\pm0.0795}$ & $0.4321_{\pm0.0453}$ & $0.2045_{\pm0.0446}$ & $0.1211_{\pm0.0262}$ & $\underline{0.0755_{\pm0.0270}}$ & $0.2316_{\pm0.0399}$ & $\mathbf{0.0504_{\pm0.0341}}$ \\
& Peer & $0.5170_{\pm0.0360}$ & $0.4451_{\pm0.0574}$ & $0.5657_{\pm0.0263}$ & $\underline{0.0543_{\pm0.0169}}$ & $0.0547_{\pm0.0079}$ & $0.0980_{\pm0.0310}$ & $0.1005_{\pm0.0255}$ & $\mathbf{0.0219_{\pm0.0088}}$ \\
& Total & $0.7625_{\pm0.0767}$ & $0.6205_{\pm0.0954}$ & $0.7070_{\pm0.0691}$ & $0.2813_{\pm0.0643}$ & $0.1930_{\pm0.0314}$ & $\underline{0.1203_{\pm0.0395}}$ & $0.4400_{\pm0.0769}$ & $\mathbf{0.0531_{\pm0.0201}}$ \\
\bottomrule
\end{tabular}}
\end{table*}

\begin{table*}[!ht]
\centering
\caption{Causal effect estimation on the \texttt{hete\_z} datasets measured by $\sqrt{\epsilon_{\mathrm{PEHE}}}$.}
\label{tab:hetero_hete_z_effect}
\resizebox{\textwidth}{!}{
\begin{tabular}{clllllllll}
\toprule
Datasets & Effects & TARNet+Z & CFR+Z & NetDeconf+Z & NetEst & RRNet & TNet$^\dagger$ & gDIS & NEDNet \\
\midrule
\multirow{3}{*}{\begin{tabular}[c]{@{}c@{}}BC\\(Within-sample)\end{tabular}}
& Ind. & $0.3758_{\pm0.0896}$ & $0.2483_{\pm0.0770}$ & $0.4571_{\pm0.0672}$ & $0.1583_{\pm0.0446}$ & $0.1846_{\pm0.0506}$ & $\underline{0.0826_{\pm0.0353}}$ & $0.2327_{\pm0.0925}$ & $\mathbf{0.0483_{\pm0.0308}}$ \\
& Peer & $0.8738_{\pm0.0907}$ & $0.7210_{\pm0.1095}$ & $0.8763_{\pm0.1040}$ & $0.1180_{\pm0.0276}$ & $\underline{0.0665_{\pm0.0106}}$ & $0.1408_{\pm0.0367}$ & $0.1881_{\pm0.0523}$ & $\mathbf{0.0316_{\pm0.0072}}$ \\
& Total & $1.3591_{\pm0.2050}$ & $0.9243_{\pm0.1675}$ & $1.1592_{\pm0.2410}$ & $0.3703_{\pm0.0922}$ & $0.3486_{\pm0.0995}$ & $\underline{0.1793_{\pm0.0669}}$ & $0.4843_{\pm0.1379}$ & $\mathbf{0.1155_{\pm0.0242}}$ \\
\midrule
\multirow{3}{*}{\begin{tabular}[c]{@{}c@{}}BC\\(Out-of-sample)\end{tabular}}
& Ind. & $0.3837_{\pm0.0943}$ & $0.2448_{\pm0.0753}$ & $0.4510_{\pm0.0618}$ & $0.1602_{\pm0.0480}$ & $0.1841_{\pm0.0504}$ & $\underline{0.0849_{\pm0.0360}}$ & $0.2317_{\pm0.0936}$ & $\mathbf{0.0516_{\pm0.0329}}$ \\
& Peer & $0.8722_{\pm0.0913}$ & $0.7201_{\pm0.1103}$ & $0.8753_{\pm0.1044}$ & $0.1169_{\pm0.0323}$ & $\underline{0.0655_{\pm0.0100}}$ & $0.1437_{\pm0.0369}$ & $0.1885_{\pm0.0513}$ & $\mathbf{0.0321_{\pm0.0065}}$ \\
& Total & $1.3453_{\pm0.2103}$ & $0.9179_{\pm0.1658}$ & $1.1471_{\pm0.2358}$ & $0.3682_{\pm0.0949}$ & $0.3493_{\pm0.1035}$ & $\underline{0.1794_{\pm0.0739}}$ & $0.4887_{\pm0.1423}$ & $\mathbf{0.1196_{\pm0.0242}}$ \\
\midrule
\multirow{3}{*}{\begin{tabular}[c]{@{}c@{}}Flickr\\(Within-sample)\end{tabular}}
& Ind. & $0.4175_{\pm0.1491}$ & $0.2877_{\pm0.0998}$ & $0.5254_{\pm0.1504}$ & $0.2417_{\pm0.1278}$ & $0.1586_{\pm0.1025}$ & $\underline{0.1271_{\pm0.0376}}$ & $0.3137_{\pm0.0972}$ & $\mathbf{0.0800_{\pm0.0717}}$ \\
& Peer & $0.8693_{\pm0.0270}$ & $0.6870_{\pm0.0255}$ & $0.8747_{\pm0.0320}$ & $0.1199_{\pm0.0791}$ & $\underline{0.0793_{\pm0.0264}}$ & $0.1886_{\pm0.0884}$ & $0.0813_{\pm0.0218}$ & $\mathbf{0.0285_{\pm0.0038}}$ \\
& Total & $1.4266_{\pm0.2525}$ & $1.0013_{\pm0.2087}$ & $1.2466_{\pm0.2330}$ & $0.4668_{\pm0.2141}$ & $0.2812_{\pm0.1327}$ & $\underline{0.2576_{\pm0.0354}}$ & $0.4690_{\pm0.0815}$ & $\mathbf{0.1307_{\pm0.0516}}$ \\
\midrule
\multirow{3}{*}{\begin{tabular}[c]{@{}c@{}}Flickr\\(Out-of-sample)\end{tabular}}
& Ind. & $0.4088_{\pm0.1543}$ & $0.2866_{\pm0.0972}$ & $0.4974_{\pm0.1418}$ & $0.2305_{\pm0.1250}$ & $0.1601_{\pm0.1019}$ & $\underline{0.1259_{\pm0.0373}}$ & $0.3082_{\pm0.0927}$ & $\mathbf{0.0797_{\pm0.0734}}$ \\
& Peer & $0.8520_{\pm0.0219}$ & $0.6859_{\pm0.0222}$ & $0.8731_{\pm0.0311}$ & $0.1023_{\pm0.0677}$ & $\underline{0.0812_{\pm0.0226}}$ & $0.1895_{\pm0.0880}$ & $0.0845_{\pm0.0227}$ & $\mathbf{0.0309_{\pm0.0038}}$ \\
& Total & $1.3424_{\pm0.2534}$ & $0.9979_{\pm0.1980}$ & $1.2119_{\pm0.2132}$ & $0.4186_{\pm0.1803}$ & $0.2813_{\pm0.1225}$ & $\underline{0.2595_{\pm0.0404}}$ & $0.4573_{\pm0.0770}$ & $\mathbf{0.1330_{\pm0.0528}}$ \\
\bottomrule
\end{tabular}}
\end{table*}

\subsection{Effect Estimation under Heterogeneous Responses}
\label{sec:heterogeneous_effects}

We further evaluate NEDNet under two heterogeneous outcome-generation settings adopted from TNet~\cite{chen2024doubly}. In the \texttt{hete} setting, individual-treatment effects vary across nodes through covariate-dependent response terms, while neighborhood-exposure effects remain homogeneous. The \texttt{hete\_z} setting further introduces node-dependent heterogeneity in neighborhood-exposure effects. The corresponding results are reported in Tables~\ref{tab:hetero_hete_effect} and~\ref{tab:hetero_hete_z_effect}, respectively. Directly transferring the hyperparameter configuration from homogeneous to heterogeneous settings led to unstable optimization for TNet. We therefore selected its hyperparameters separately for each heterogeneous setting using validation data only. The resulting variants are denoted by TNet$^\dagger$ in the tables.

Table~\ref{tab:hetero_hete_effect} shows that NEDNet achieves the lowest PEHE across all twelve dataset--evaluation--effect combinations. Compared with the best-performing baseline in each setting, NEDNet reduces the estimation error by 33.2\%--59.7\%. The improvements are consistent across both datasets and evaluation settings, covering individual, peer, and total effects. These results demonstrate the effectiveness of NEDNet in estimating heterogeneous individual-treatment effects while maintaining accurate estimation of neighborhood-exposure and total effects.

The results on the \texttt{hete\_z} setting further examine the estimation performance when both individual-treatment and neighborhood-exposure effects exhibit node-dependent heterogeneity. As reported in Table~\ref{tab:hetero_hete_z_effect}, NEDNet again achieves the lowest PEHE in all twelve evaluation combinations, reducing the error by 33.3\%--64.1\% relative to the best-performing baseline. The consistent improvements across IE, PE, and TE indicate that NEDNet can effectively estimate heterogeneous responses along both intervention dimensions. Moreover, the comparable within-sample and out-of-sample results suggest that its estimation performance is maintained on held-out graph subgraphs.

\section{Conclusion}
\label{sec:conclusion}

This paper has presented NEDNet, a structured conditional outcome model for causal effect estimation under network interference. NEDNet has parameterized the conditional outcome response using a query-invariant pre-treatment baseline and separate pathways for individual-treatment and neighborhood-exposure responses. The path-specific neighborhood contextualization mechanism has further incorporated local pre-treatment information into the two pathways. This design has supported potential-outcome prediction under alternative treatment--exposure conditions and the estimation of individual, peer, and total effects.

NEDNet has been evaluated on semi-synthetic benchmarks derived from BlogCatalog and Flickr under homogeneous and covariate-heterogeneous response settings. Across the evaluated comparisons, NEDNet has achieved the lowest mean PEHE among the compared methods for IE, PE, and TE, and has maintained competitive counterfactual outcome prediction performance. The ablation results have further supported the effectiveness of separate response representations and path-specific encoding and decoding.

Future work will investigate more general response specifications, including joint treatment--exposure interactions, and more flexible exposure mappings to accommodate increasingly complex network interference patterns.

\bibliographystyle{IEEEtran}
\bibliography{TNNLS}

\end{document}